\documentclass{article}
\usepackage{spconf,amsmath,graphicx}
\usepackage{multirow}
\usepackage{enumitem}
\setlist{nosep, leftmargin=14pt}
\usepackage{mwe}
\usepackage[draft]{hyperref}
\hypersetup{
  bookmarks=false,
  pdfborder={0 0 0}
}

\usepackage[table]{xcolor} 
\usepackage{cite}
\usepackage{etoolbox}
\apptocmd{\thebibliography}{
    \setlength{\itemsep}{0pt}
    \setlength{\parskip}{0pt}
}{}{}

\title{Joint Segmentation and Grading with Iterative Optimization for Multimodal Glaucoma Diagnosis}

\name{Zhiwei Wang\textsuperscript{1,2},
Yuxing Li\textsuperscript{1},
Meilu Zhu\textsuperscript{1},
Defeng He\textsuperscript{2},
Edmund Y. Lam\textsuperscript{1,*}
\thanks{Corresponding author: Edmund Y. Lam (elam@eee.hku.hk)}
\thanks{This work is supported by the Innovation and Technology Fund (ITS/341/23) of Hong Kong, China.}}
\address{
\textsuperscript{1}Department of Electrical and Electronic Engineering, The University of Hong Kong, Hong Kong, China\\
\textsuperscript{2}College of Information Engineering, Zhejiang University of Technology, Hangzhou, China
}

\begin{document}
\maketitle

\begin{abstract}
Accurate diagnosis of glaucoma is challenging, as early-stage changes are subtle and often lack clear structural or appearance cues. Most existing approaches rely on a single modality, such as fundus or optical coherence tomography (OCT), capturing only partial pathological information and often missing early disease progression.
In this paper, we propose an iterative multimodal optimization model (IMO) for joint segmentation and grading. IMO integrates fundus and OCT features through a mid-level fusion strategy, enhanced by a cross-modal feature alignment (CMFA) module to reduce modality discrepancies. An iterative refinement decoder progressively optimizes the multimodal features through a denoising diffusion mechanism, enabling fine-grained segmentation of the optic disc and cup while supporting accurate glaucoma grading.
 Extensive experiments show that our method effectively integrates multimodal features, providing a comprehensive and clinically significant approach to glaucoma assessment.
Source codes are available at \href{https://github.com/warren-wzw/IMO.git}{https://github.com/warren-wzw/IMO.git}.
\end{abstract}

\begin{keywords}
Glaucoma, Iterative optimization, multimodal data, Segmentation, Grading
\end{keywords}
\begin{figure*}[t]
    \centering
    \includegraphics[width=0.95\textwidth]{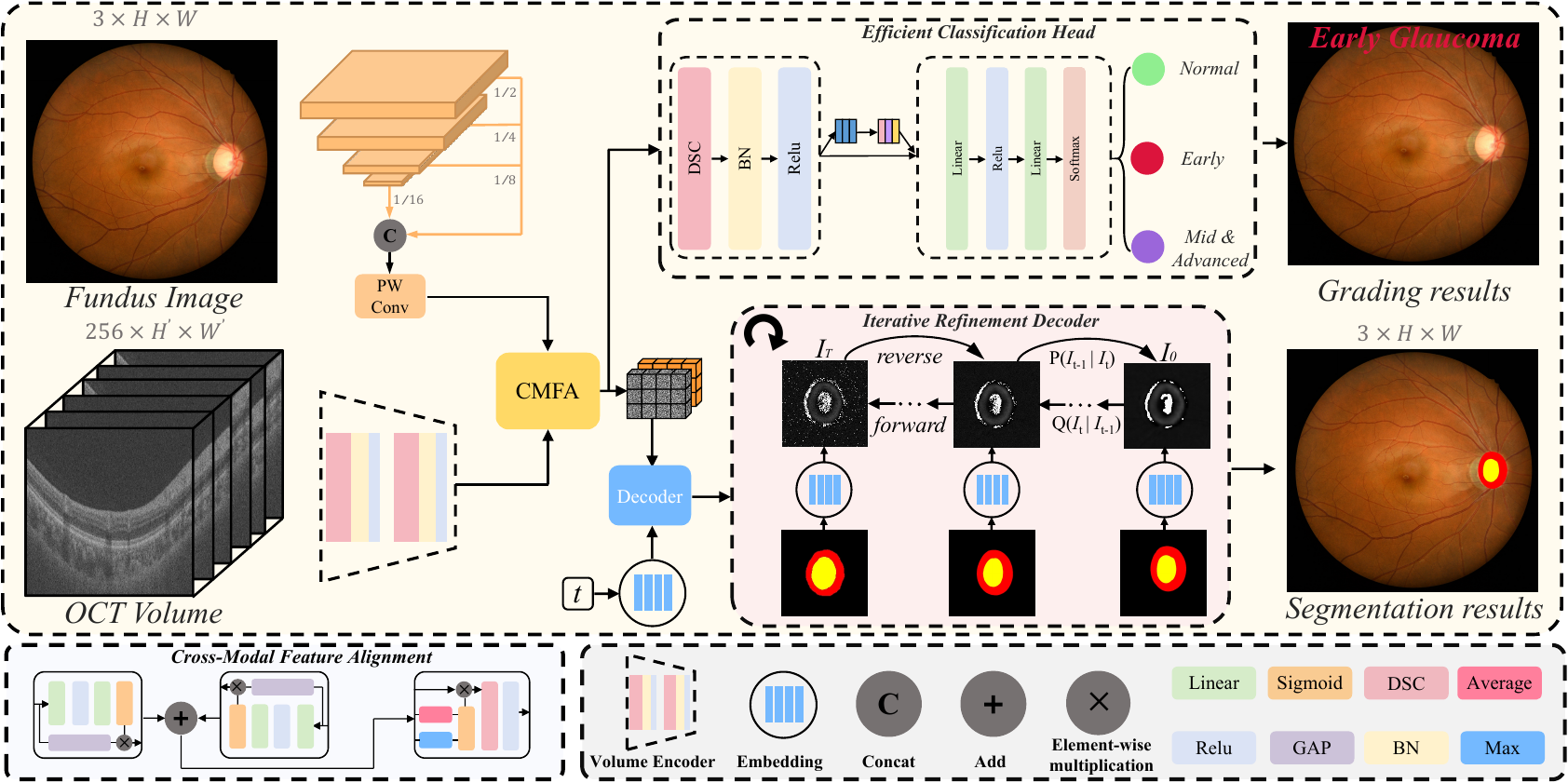}
    \caption{The architecture of the proposed iterative multimodal optimization model.}
    \vspace{-0.3cm}
    \label{ModelArch}
\end{figure*}

\section{Introduction}
\label{sec1}
High ocular disease prevalence has made automated diagnosis a key research focus \cite{li2024intelligent,Tham2014GlobalPO}. Despite progress, accurate glaucoma diagnosis remains challenging due to subtle structural and morphological changes \cite{Li2023ADM}. Therefore, developing precise and clinically meaningful assessment methods remains a vital yet difficult task.

Various single-modality segmentation \cite{Zhu2023FedDMFW,Cho2024EnhancedSP} and grading \cite{Liang2024LAGNetLA} methods have been proposed for glaucoma diagnosis using fundus or OCT data. For instance, Liang et al. \cite{Liang2024LAGNetLA} utilized a vision transformer and graph neural network for grading. Cho et al. \cite{Cho2024EnhancedSP} developed an unsupervised domain adaptation framework for segmentation. To overcome the inherent limitations of single-modality data, multimodal strategies combining fundus and OCT have emerged \cite{Cai2022CorollaAE, Li2024DualModeIS}. Cai et al. \cite{Cai2022CorollaAE} converted OCT volumes into retinal thickness maps to complement fundus images, enhancing grading. However, current multimodal methods treat segmentation and grading as independent tasks, failing to exploit their inherent correlation and the potential for joint optimization.

To more effectively leverage multimodal complementarities and harness the potential of joint multi-task optimization \cite{Zhu2021DSINetDS, Wang2025DiFusionSegDS}, we propose IMO, a novel model for glaucoma diagnosis. The framework processes fundus and OCT data, adopting mid-level fusion with dual encoders to extract features. To address modality discrepancies, a CMFA module aligns representations in the shared feature space. An efficient grading head and iterative segmentation decoder are also introduced to progressively refine results. Through stepwise iterative optimization, IMO jointly optimizes segmentation and grading, achieving 93.33\% mDice and 81.48\% precision on the GAMMA dataset, demonstrating the effectiveness of multi-task optimization. 
The main contributions of this paper can be summarized as follows:
\begin{itemize}
\item We propose IMO, a joint framework integrating fundus and OCT images for concurrent glaucoma grading and optic disc/cup segmentation.

\item We develop a CMFA module for cross-modality feature alignment and a diffusion-based decoder for iterative, progressive segmentation refinement.
\end{itemize}

\section{PROPOSED METHOD}\label{sec2}
\subsection{Overall Framework}
\autoref{ModelArch} illustrates the framework. Mid-level fusion integrates fundus and OCT features via dual encoders for modality-specific representations.

A feature pyramid encoder extracts multi-scale fundus features using progressive downsampling and pointwise convolution. Simultaneously, a network captures complementary volumetric OCT cues. The CMFA module aligns these encoder outputs into a unified representation, integrating structural and appearance cues. These fused features are then fed into two parallel task-specific branches: an iterative refinement decoder for optic disc/cup segmentation and an efficient classification head.

The classification branch predicts glaucoma stages from fused features. Depthwise separable convolutions (DSC), batch normalization, and ReLU efficiently extract spatial-channel information, followed by a SE block to model channel dependencies. Finally, a two-layer fully connected head and softmax layer output stage probabilities.

\subsection{Cross-Modal Feature Alignment Module}
Fundus and OCT volumes differ, leading to misaligned features. To enhance fused representations, we design CMFA to align and improve expressiveness, as shown in \autoref{ModelArch}.

Specifically, CMFA first applies channel attention to each modality. Global Average Pooling (GAP) captures global channel statistics, and a two-layer fully connected network generates channel-wise attention weights \(W_1\) and \(W_2\) to emphasize informative channels and suppress redundant ones. This captures modality-specific differences along the channel dimension, ensuring that important features are prioritized during fusion. The operation is formulated as follows:
\begin{equation}
    \begin{aligned}
       &X_{\mathrm{att}}=X\odot\sigma(W_2\cdot{ReLU}(W_1\cdot{GAP}(X))), \\
    \end{aligned}
\end{equation}
where \(X\) denotes the input feature map, \(\sigma\) denotes the sigmoid function and \(\odot\) means element-wise multiplication. 

The channel-refined features are then summed to form an initial fused representation. To further enhance spatial structural information, a spatial attention mechanism is introduced. By computing channel-wise average and max pooling, a spatial attention map \(S=\sigma({Conv}({Mean}(X_{att}^{f+o})\oplus{Max}(X_{att}^{f+o}))\) is generated to highlight the salient regions in the feature map. This step emphasizes key spatial regions across modalities and can be expressed as:
\begin{equation}
    \begin{aligned}
       & F_{{fused}}^{spatial}=X_{att}^{f+o}\odot S.\\
    \end{aligned}
\end{equation}

Finally, a pointwise convolution followed by ReLU activation integrates channel information and introduces non-linearity, yielding the final fused feature. This output contains complementary information from both modalities and benefits from refined channel and spatial selection, providing a stronger representation for downstream tasks:
\begin{equation}
    \begin{aligned}
       &F_{{\mathrm{out}}}=ReLU(Conv_{1\times1}(F_{{{fused}}}^{spatial})). \\
    \end{aligned}
\end{equation}

\subsection{Iterative Refinement Decoder}
Traditional segmentation methods often adopt a single-step segmentation paradigm, which is simple and efficient but typically struggles in complex multimodal segmentation scenarios. Inspired by the success of DDPM \cite{Ho2020DenoisingDP}, which progressively refines predictions through multistep iterations, we designed an iterative optimization decoder for the multimodal segmentation task. Its overall architecture is illustrated in \autoref{ModelArch}. In the forward process, we follow the standard noise injection scheme of DDPM:
\begin{equation}
    \begin{aligned}
    X_t=\sqrt{\overline{\alpha}_t}X_0+\sqrt{1-\overline{\alpha}_t}\epsilon ,
    \end{aligned}
\end{equation}
where \(\bar{\alpha}_T=\prod_{i=1}^T\alpha_i\), and \(\alpha_i\) controls the intensity of the noise at each time step.

During the reverse denoising phase, we adopt the DDIM sampling strategy to accelerate the inference process. The sampling process can be expressed as:
\begin{equation}
    \begin{aligned}
    x_{t-n}=\sqrt{\overline\alpha_{t-n}}\hat{\mathbf{x}}_0+\sqrt{1-\overline{\alpha}_{t-n}}\epsilon_\theta(\mathbf{x}_t,t), 
    \end{aligned}
\end{equation}
where \(\hat{\mathbf{x}}_0\) denotes the clean segmentation representation predicted in the timestep \(t\), \(\overline{\alpha}_{t-n}\) is the cumulative product of the noise scheduling coefficients controlling the denoising trajectory, and \(\epsilon_\theta(\mathbf{x}_t,t)\) represents the noise estimated by the diffusion model at step \(t\).

Through this iterative process, the model progressively refines the segmentation mask using fused features, rather than relying on a static fundus image. This approach improves robustness to uncertainty, noise, and modality discrepancies, yielding more accurate and precise boundaries.


\section{Experiments}\label{sec3}
\subsection{Experiment setup}

\textbf{Dataset: }Experiments utilize the GAMMA dataset \cite{Wu2022GAMMACG}, containing 100 fundus/OCT pairs with glaucoma labels and disc/cup masks. An 80/20 random split is used.

\noindent \textbf{Implementation Details: }Experiments are conducted in Pytorch 2.2 on an RTX 4090 GPU for \(1 \times 10^{5}\) iterations, learning rate of \(1 \times 10^{-5}\), and batch size 6.

\begin{table}[t]
    \centering
    \tiny
    \resizebox{0.48\textwidth}{!}{
        \begin{tabular}{lccccc}
            \hline 
            \multirow{2}{*}{Methods} & Unlabel & Optic Disc & Optic Cup & \multirow{2}{*}{mDice$\uparrow$} \\
            \cline{2-4}
            & Dice$\uparrow$ & Dice$\uparrow$ & Dice$\uparrow$ & \\
            \hline
            UNet++\cite{Zhou2019UNetRS} & 99.94 & 84.27 & 85.12 & 89.78 \\
            DeepLabV3Plus\cite{Chen2018EncoderDecoderWA} & 99.93 & 81.08 & 80.58 & 87.20 \\
            UPerNet\cite{Xiao2018UnifiedPP} & 99.94 & 82.85 & 84.94 & 89.24 \\
            SegFormer\cite{Xie2021SegFormerSA} & 99.95 & \underline{86.18} & \underline{86.31} & \underline{90.81} \\
            SegNext\cite{Guo2022SegNeXtRC}  & \underline{99.96} & 85.84 & 85.92 & 90.57 \\
            Ours                     & \textbf{99.97} & \textbf{89.93} & \textbf{90.09} & \textbf{93.33} \\
            \hline
        \end{tabular}
    }
    \caption{Segmentation results of different methods on our dataset. The best and second-best results are highlighted in \textbf{bold} and \underline{underlined}.}
    \label{SegTable}
    \vspace{-0.3cm}
\end{table}

\begin{figure}[t]
    \centering
    \includegraphics[width=0.46\textwidth]{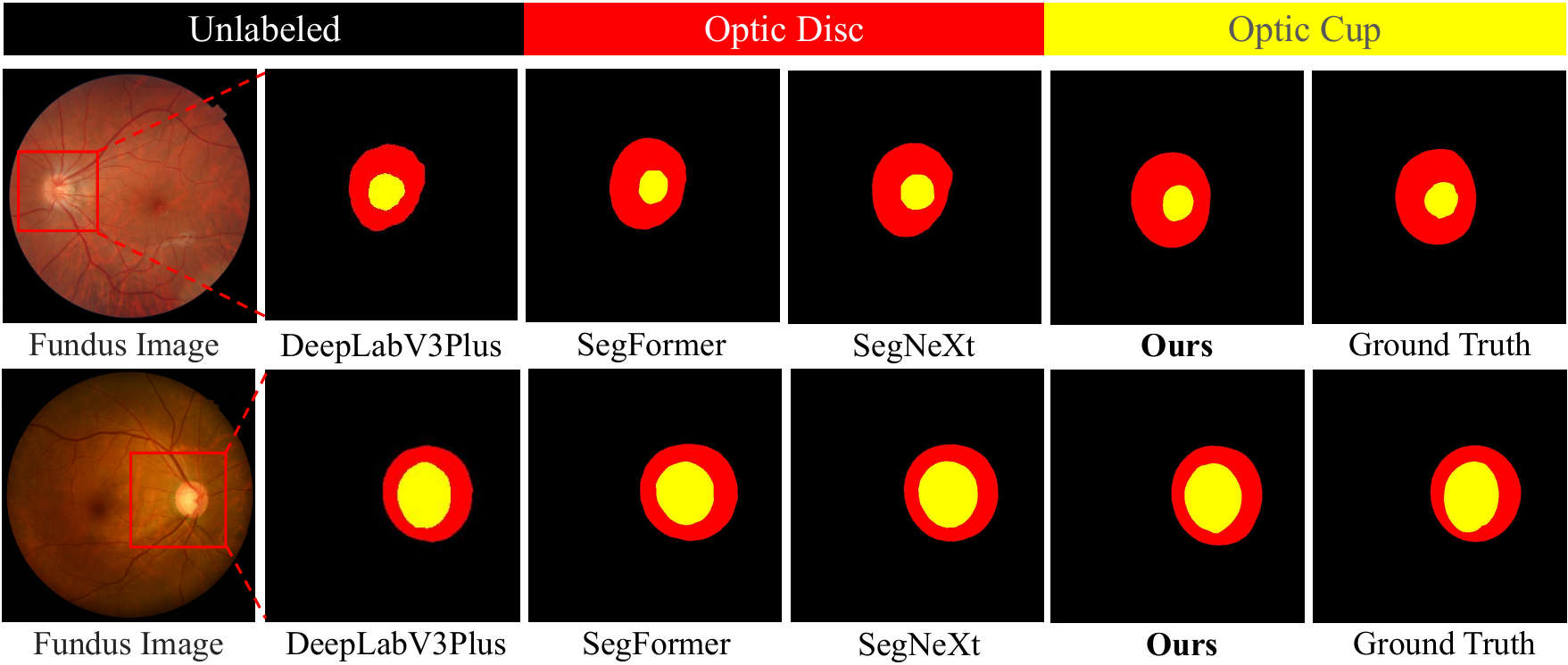}
    \caption{The segmentation results are shown for each scene in the rows. To better visualize the results, the important regions of each image have been enlarged.}
    \label{SegResult}
    \vspace{-0.3cm}
\end{figure}
\subsection{Comparison and Analysis}
We compare IMO against several segmentation and classification baselines. As shown in \autoref{SegTable}, IMO outperforms other methods in segmentation. Specifically, it achieves 89.93\% and 90.09\% Dice for the optic disc and cup, respectively, notably surpassing second-best results. Overall, our approach reaches the highest mean Dice of 93.33\% on the test set. These results indicate that IMO improves both pixel-level overlap and boundary delineation, which are essential for subsequent glaucoma diagnosis.

To better illustrate the comparison, we provide the visual results of our method alongside other approaches, as shown in \autoref{SegResult}. For the first image, our segmentation output is noticeably smoother and more precise than that of the alternative method, with the optic cup region showing natural boundaries and consistent structural details. In the second image, our method achieves higher accuracy and finer delineation, capturing subtle features that are missed or incorrectly segmented by other approaches. In general, the visual comparison demonstrates the superior ability of our method to produce clinically accurate smooth segmentation maps.

\autoref{ClsTable} presents a quantitative comparison of glaucoma grading across different methods. Our approach achieves the highest recall of 79.17\% and F1-score of 72.03\%, demonstrating strong capability to correctly identify positive cases while maintaining balanced performance. Our precision reaches 81.48\%, slightly below ResNet-50 by 0.74\%, while our method ties for the highest accuracy of 75.00\%. These results indicate that our approach not only performs well in segmentation, but also excels in grading tasks.

\subsection{Ablation Study}
\noindent \textbf{Framework Ablation: }To evaluate the effectiveness of our joint optimization framework, we reconstructed several variant configurations, including single-modality input segmentation-only, and grading-only. As shown in the first three rows of \autoref{AblationStudy}, using only fundus images (w/o OCT) leads to a decrease in all metrics, with grading Precision decreasing by 11.36\%. Optimizing a single task does not outperform joint optimization, demonstrating that our framework effectively leverages the complementary information between segmentation and grading.

\noindent \textbf{CMFA Ablation: }To assess the contribution of the CMFA, we remove CMFA and instead directly concatenate features from both modalities. As shown in the fourth row of \autoref{AblationStudy}, this degradation leads to unstable feature interaction and weakens the modality complementarity. Quantitatively, mDice from 93.33\% to 92.72\%, and grading Precision drops by 7.24\%, indicating that CMFA effectively aligns the features from different modalities and makes it suitable for downstream tasks.

\noindent \textbf{Iterative Refinement Decoder Ablation: }As referred to in the fifth row of \autoref{AblationStudy}, replacing the iterative refinement decoder (IRD) with a single-pass decoder leads to a decrease in mDice of about 0.47\%, the impact is relatively minor, with Precision decreasing by only 1.09\% relative to the other ablation configurations. This confirms that IRD facilitates progressive feature refinement, leading to more stable segmentation boundaries and improved grade reliability.

\begin{table}[t]
    \centering
    \resizebox{0.46\textwidth}{!}{ 
        \begin{tabular}{lcccc}
            \hline
            Methods & Precision$\uparrow$  & Accuracy$\uparrow$ & Recall$\uparrow$ & F1-score$\uparrow$ \\
            \hline
            ResNet-50 \cite{He2015DeepRL}      & \textbf{82.22} & \textbf{75.00} & 66.67 & 66.49 \\
            ConvNeXt \cite{Liu2022ACF}      & 79.05 & 65.00 & 58.33 & 58.09 \\
            ViT \cite{Dosovitskiy2020AnII}            & 77.78 & \underline{70.00} & 66.67 & \underline{67.94} \\
            BEiTv2 \cite{Peng2022BEiTVM}        & 72.22 & 65.00 & 66.67 & 60.00 \\
            EfficientViT \cite{Liu2023EfficientViTME}   & 74.24 & 70.00 & \underline{70.83} & 66.25 \\
            Ours  & \underline{81.48} & \textbf{75.00} & \textbf{79.17} & \textbf{72.03} \\
            \hline
        \end{tabular}
    }
    \caption{Comparison of glaucoma grading performance across different methods.}
    \label{ClsTable}
    \vspace{-0.3cm}
\end{table}

\begin{table}[t]
    \centering
    \resizebox{0.46\textwidth}{!}{ 
        \begin{tabular}{lcccc}
        \hline
        Method & Disc (Dice$\uparrow$) & Cup (Dice$\uparrow$) & mDice$\uparrow$ & Precision$\uparrow$ \\
        \hline
        w/o OCT   & 87.38 & 88.02 & 91.79 & 70.12 \\
        w/o Grd.  & 88.93 & 89.51 & 92.80 & -     \\
        w/o Seg.  & -     & -     & -     & 78.41 \\
        w/o CMFA  & \underline{89.22} & 88.98 & 92.72 & 74.24 \\
        w/o IRD   & 89.03 & \underline{89.59} & \underline{92.86} & \underline{80.39} \\
        Ours      & \textbf{89.93} & \textbf{90.09} & \textbf{93.33} & \textbf{81.48} \\
        \hline
        \end{tabular}
    }
    \caption{Ablation studies of IMO. ``w/o'' denotes without, while ``Grd.'' and ``Seg.'' refer to the grading and segmentation tasks, respectively.}
    \label{AblationStudy}
    \vspace{-0.3cm}
\end{table}
\section{Conclusion}\label{sec4}
In this paper, we propose the IMO, which leverages complementary information from fundus images and OCT volumes for joint segmentation and grading. The iterative multimodal framework enables mutual supervision between the two tasks, enhancing feature integration and overall performance. Experiments on the GAMMA dataset demonstrate strong results in both segmentation and grading. The potential of the framework can be further unlocked with more diverse and well-annotated multimodal data. 

\section{COMPLIANCE WITH ETHICAL STANDARDS}\label{sec6}
The use of open-access datasets complied with their license terms and did not require separate ethical approval.

\bibliographystyle{IEEEtran}
\bibliography{refs_4pages}

\end{document}